\documentclass[conference]{IEEEtran}
\IEEEoverridecommandlockouts
\usepackage{cite}
\usepackage{amsmath,amssymb,amsfonts}
\usepackage{easyReview}
\usepackage{algorithmic}
\usepackage{float}
\usepackage{graphicx}
\graphicspath{ {./images/} }
\usepackage{textcomp}
\usepackage{xcolor}
\usepackage[ruled,vlined]{algorithm2e}
\usepackage{subfig}

\def\BibTeX{{\rm B\kern-.05em{\sc i\kern-.025em b}\kern-.08em
    T\kern-.1667em\lower.7ex\hbox{E}\kern-.125emX}}
    
\begin{document}
\bstctlcite{IEEEexample:BSTcontrol}

\title{Enabling Multi-Agent Transfer Reinforcement Learning via Scenario Independent Representation}

\author{

    \IEEEauthorblockN{1\textsuperscript{st} Ayesha Siddika Nipu}
    \IEEEauthorblockA{\textit{Department of Computer Science} \\
    \textit{Missouri State University}\\
    Springfield, U.S. \\
    Nipu62@MissouriState.edu}
    
    \and
    \IEEEauthorblockN{2\textsuperscript{nd} Siming Liu}
    \IEEEauthorblockA{\textit{Department of Computer Science} \\
    \textit{Missouri State University}\\
    Springfield, U.S. \\
    SimingLiu@MissouriState.edu}
    
    \and
    \IEEEauthorblockN{3\textsuperscript{rd} Anthony Harris}
    \IEEEauthorblockA{\textit{Department of Computer Science} \\
    \textit{Missouri State University}\\
    Springfield, U.S. \\
    Anthony999@MissouriState.edu}
}




\maketitle 

\IEEEpubidadjcol
\begin{abstract} 
Multi-Agent Reinforcement Learning (MARL) algorithms are widely adopted in tackling complex tasks that require collaboration and competition among agents in dynamic Multi-Agent Systems (MAS). 
However, learning such tasks from scratch is arduous and may not always be feasible, particularly for MASs with a large number of interactive agents due to the extensive sample complexity. 
Therefore, reusing knowledge gained from past experiences or other agents could efficiently accelerate the learning process and upscale MARL algorithms. 
In this study, we introduce a novel framework that enables transfer learning for MARL through unifying various state spaces into fixed-size inputs that allow one unified deep-learning policy viable in different scenarios within a MAS. 
We evaluated our approach in a range of scenarios within the StarCraft Multi-Agent Challenge (SMAC) environment, and the findings show significant enhancements in multi-agent learning performance using maneuvering skills learned from other scenarios compared to agents learning from scratch.
Furthermore, we adopted Curriculum Transfer Learning (CTL), enabling our deep learning policy to progressively acquire knowledge and skills across pre-designed homogeneous learning scenarios organized by difficulty levels. 
This process promotes inter- and intra-agent knowledge transfer, leading to high multi-agent learning performance in more complicated heterogeneous scenarios.

\end{abstract}

\begin{IEEEkeywords}
Deep reinforcement learning, multi-agent system, transfer learning, curriculum learning, StarCraft II
\end{IEEEkeywords}

\section{Introduction}\label{lab-intro}

Significant achievements have been made in the field of Artificial Intelligence (AI) in the past decade. Gaming environments, such as Atari video games \cite{mnih2015human}, board games \cite{silver2016mastering, shao2016move}, poker games \cite{moravvcik2017deepstack}, and autonomous driving simulations\cite{muhammad2020deep}, have served as valuable testing grounds for AI research. Such games offer fixed, finite action spaces and a single-agent environment, simplifying the learning objectives for AI researchers. However, many real-world problems that feature complex rules and multiple agents pose greater challenges for AI research. Considering the involvement of various agents in cooperative and competitive applications, we strive to explore AI techniques for  Multi-Agent Systems (MAS) beyond single-agent systems.
Among many popular AI approaches, reinforcement learning (RL) in combination with deep neural networks (DNNs) has gathered significant attention to tackle real-world problems in recent years \cite{schmidhuber2015deep, littman2015reinforcement, mnih2015human}. The framework of deep RL (DRL) offers a promising approach for enabling intelligent agents to learn end-to-end solutions to complex tasks as competitively as human-level experts in many fields. The deep Q-network (DQN) technique leverages the experience replay method and a target network to mitigate the correlation between samples, and thus, the process of training is stabilized \cite{mnih2015human}. As an example, AlphaGo \cite{wang2016does}, a computer program that defeated the world champion in the board game Go, utilizes a variant of DRL, known as the policy gradient method. However, expanding single-agent RL algorithms to solve MAS problems is not straightforward.
Multi-agent reinforcement learning (MARL) poses a major challenge due to the limited generalization capability of conventional RL algorithms, such as Q-learning and policy gradient methods. This issue is caused by the exponential growth in the number of states with an increasing number of agents. To address these challenges and accomplish team objectives, a popular approach emerged utilizing centralized training and decentralized execution (CTDE), where the complete information is used to train agents for MARL.

Although DRL and MARL have made tremendous successes in many fields, the enormous amount of training samples and extensive learning duration significantly limit their applicability to complex problems, especially in large scale multi-agent settings.  
To alleviate sample complexity and accelerate the learning process of autonomous agents in tackling complex learning tasks, transfer learning (TL) has attracted much attention from researchers. 
TL proposes to reuse knowledge from previous tasks or external sources, such as demonstrations from humans or advice from other learning agents, to speed up the learning process. Unprincipled reuse of knowledge, however, can lead to negative transfer, making learning more difficult. 
To develop flexible and robust methods for autonomously reusing knowledge, TL for single-agent RL has evolved significantly to be usable in complex applications. Nevertheless, multi-agent TL approaches still require further development to find real-world applications and achieve autonomous learning. 

Significant efforts have been made in developing dedicated neural networks (NN) and comprehensive training techniques to enable TL in multi-agent settings \cite{da2019survey}. 
Efficiently learning policies from scratch in multi-agent settings is difficult and time-consuming, particularly for tasks with exploration challenges. In such settings, how to unify multi-modal data encoding and decoding to enable agents’ knowledge transferring and curriculum learning is essential to increase MARL performance.
One attempt to enable TL for MADDPG has been made by Zhang et al. in \cite{zhang2019efficient} for training agents in multi-UAV combat. In that study, the input dimension of the network structure is proportional to the number of agents, representing the agent's observation of the entire environment. However, this approach can only be applied for transfer training when the number of agents is consistent. 
The question of enabling autonomous agents that can learn faster by reusing knowledge from various sources in MAS remains open.

In order to address the aforementioned issues and overcome the existing challenges, we propose a novel spatial and feature encoding framework for unifying the state inputs of individual agents to the neural network along with a generalized output representation regardless of different multi-agent scenarios.
We utilize a spatial abstraction technique Influence Map (IM), to unify various local observations into a fixed dimension in combination with abstracted global information using Multi-Agent Influence Map (MAIM) \cite{harris2021maidrl} for agents to achieve scenario-independent capability.
This spatial and feature representation is aggregated to an agent's previous and current states and trained with fixed-size neural network policies preserving domain knowledge across multiple scenarios in MAS. 
We then conducted a rigorous analysis to evaluate the performance of our TL model in different StarCraft Multi-Agent Challenge (SMAC) scenarios. Our approach shows promising results regarding robustness and scalability among agents for intra- and inter-agent knowledge transfer.

The remainder of this paper is partitioned as follows: 
Section~\ref{relWork} explores the related works in MARL and TL.
Section~\ref{methodology} explains the methodology regarding our experimentation. The results from the experiments have been demonstrated in Section~\ref{results} afterward, and Section~\ref{conclusion} draws the conclusion and suggests the future scope of this work.

\section{Related Work}\label{relWork}

Numerous research has been performed using RL algorithms to train collaborative agents to achieve team goals in MAS environments. 
Konda et al. presented the actor-critic (AC) algorithm that integrated value-based and policy-based learning approaches by utilizing both Q-Learning and policy gradient approaches \cite{konda1999actor}.
Schulman et al. introduced Proximal Policy Optimization (PPO), and this method has shown the ability to effectively reduce the variance in the results \cite{Schulman2017}.
Yu et al. extended the PPO into Multi-Agent PPO (MAPPO), specializing in multi-agent settings \cite{yu2021surprising}.

\begin{figure}[tb]
    \centering
    \includegraphics[width=0.35\textwidth]{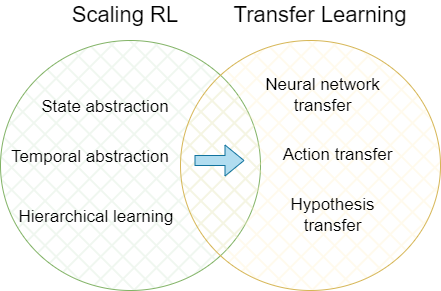}
    \caption{Transfer Between Reinforcement Learning Tasks }
    \label{fig:TransferLearning}
    \vspace{-6mm}
\end{figure}

In MAS with large states, the decision for action-space becomes challenging due to the complex learning policies. Many researchers worked on resolving numerous issues by exploiting information from other training models. 
Fig.~\ref{fig:TransferLearning} provides an overall insight into transferring knowledge between RL methods. The arrow-pointed portion denotes the strategy to follow for transferring such information. 
In a recent survey, Silva et al. mentioned a probable flaw in TL strategy as most researchers assume the knowledge during the training phase will be consistent \cite{da2019survey}. However, the assumption may not be accurate in real-world applications as the agents might not have the perfect knowledge of the task. In such cases, despite the knowledge carried from prior tasks, agents still need to explore the environment to learn the optimal policy. Koga et al. proposed to aggregate multiple policies learned from the environment into a single stochastic abstracted representation \cite{koga2014stochastic}. Though this approach performs well in multi-agent scenarios, generalizing the information sometimes lacks choosing the optimal policy, especially when the number of information is too high. This raises scalability issues that require further investigation of feature extraction before generalizing the policies.
Mahmud and Ray introduced a term, negative transfer, where they used conditional Kolmogorov complexity to find the correlation using the Bayesian framework \cite{mahmud2007transfer}. 
Taylor et al. shared that finding the optimal knowledge to transfer relies on the standard metrics tuned for the training phase \cite{taylor2009transfer}. He also extended by stating that arbitrary information might not produce the expected outcome because of wrong learning bias.
Jason et al. \cite{yosinski2014transferable} proposed a methodology to measure the transferability of characteristics at individual layers of a neural network, which sheds light on the level of generalization. 
Chen et al. proposed Net2Net technique to transfer the knowledge of the previous network by using weighted values in the input \cite{chen2015net2net}. 
While Net2Net focuses on function-preserving transformations between network specifications, our approach involves adding an additional state transformation to the input states without manipulating the neural network structures.

Xu et al. introduced a new approach in which Graph Neural Network (GNN) is combined with RL algorithms for multi-agent combat problems \cite{xu2021aggregation}. The input state representation is generalized using GNN, and the training was converged expeditiously on StarCraft. 
Khan et al. \cite{khan2021leveraging} built a transformer-based neural architecture for global state representation and build order prediction. This model resolved the bias of RNN and demonstrated the superiority of transformers using positional encoding input for the decoder. Despite the outstanding performance, this approach lacks to handle the parallel loading due to the limitation of the used dataset.
In 2015, Tan et al. proposed the Transitive Transfer Learning (TTL) framework, which involves finding a suitable intermediate domain and transferring knowledge between domains, where the effectiveness of knowledge transfer is influenced by domain difficulty and distance \cite{tan2015transitive}. 
Liu et al. \cite{liu2021efficient} presented an abstract forward model named Thought Game (TG) for transfer learning that beat the cheating level-10 AI in StarCraft by $90\%$. In contrast to this work, we focused on knowledge transferring on similar problems within the same domain.    

Shao et al. \cite{shao2018starcraft} proposed a gradient-based SARSA algorithm where the inputs of neural networks are determined by the agent’s current hitpoints, cooldown, and cumulative distance of own units in the StarCraft micromanagement system. 
The goal of our research is similar to Shao's work, except for the part where we consider both local and abstracted global information in the state space and specific move actions in the action space.
In this study, we strive to resolve the challenge by defining a uniform state representation of the local observation combined with an abstracted global state regardless of the number of agents in the environment. 
The agent’s previous and current state information is also integrated with the uniform state representation to the multi-agent training process for fine-tuned decision-making.

\section{Methodology}\label{methodology}

The foundation of our work is established upon our prior work \cite{harris2021maidrl, 9893711}, in which the proposed models' performance was evaluated on various multi-agent challenge scenarios in SMAC. We continue to use SMAC as our research platform for all the experiments evaluating multi-agent learning and transfer learning performance. The detail of the SMAC environment is described in the following section.

\begin{figure}[tb]
    \centering
    \includegraphics[width=0.35\textwidth]{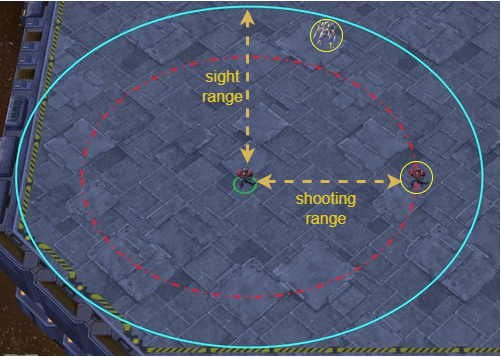}
    \caption{Single Agent's Sight Range in SMAC Scenario}
    \label{fig:sightRange}
    \vspace{-6mm}
\end{figure}

\begin{figure*}[h]
    \centering
    \includegraphics[width=1.0\textwidth]{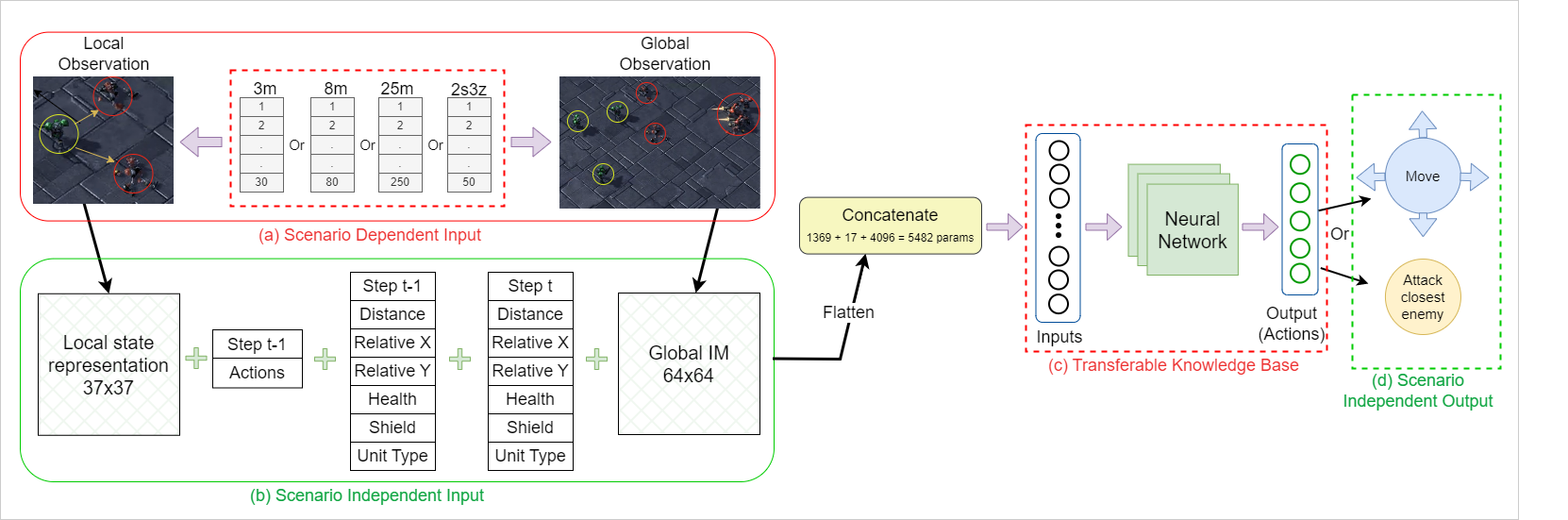}
    \caption{Transfer Learning Model Representation for Single Unit}
    \label{fig:transferLearningArc}
    \vspace{-4mm}
\end{figure*}
\subsection{Simulation Environment}
SMAC is constructed on top of the StarCraft II Learning Environment\cite{vinyals2017starcraft} and provides a wide selection of multi-agent micromanagement challenging scenarios where the goal is to eliminate opponents controlled by the built-in StarCraft AI.

Our work focuses on the micromanagement of a team of individual agents in decentralized decision-making with a shared interest, rather than the macromanagement of the army as a whole in a centralized way. 
SMAC has various homogeneous and heterogeneous scenarios, including symmetric and asymmetric distribution in each group. In each episode, SMAC provides a set of observations with state rewards and accepts a list of actions that make it ideal for RL algorithms to apply. 
The SMAC scenarios are named following a specific convention where the alphabetic character signifies the type of the agents, such as marines being abbreviated as $m$, stalkers as $s$, zealots as $z$, etc. 
The numeric value represents the number of active agents in each team. 
Our experiments utilize multiple scenarios with different difficulty levels including homogeneous scenarios $3m$, $8m$, $25m$ for evaluating intra-agent knowledge transferring and a heterogeneous scenario $2s3z$ for inter-agent knowledge reusing.

We model the SMAC scenarios as Markov games, which is a multi-agent extension of Markov Decision Processes (MDP) \cite{Peng2017, Lowe2017, Foerster2018}. A Markov game includes a set of states denoting the agents' and environment's status, and a set of actions and observations for each agent involved. $A_1,A_2,...,A_N$ denotes the actions and $O_1,O_2,...,O_N$ denotes the observations where $N$ is the number of agents in each episode.
The ally and enemy units, surrounded by the environment, are modeled for individual agents in the Markov game. The decision of action is taken by the agents' current observation.  
In a Markov game where each agent acts in accordance with a policy $\pi$ at each environmental step and earns a shared reward $r$, $r$ represents the state transition from $S$ to $S'$ ($S\times \{A_1, ..., A_N\}\rightarrow S'$).

For the scenario-independent state representation, we consider two types of observation space in our experiments. The first one is local observation, where each agent has access to only a limited range of surrounding information that the agent can observe. The other one is shared global abstraction, where all the information of the game environment is accessible from a holistic point of view. 
The local observation of an agent contains a list of properties including hitpoints, unit type, relative positions, and distance for both allied and enemy agents within the observation range. This information is received from SMAC without customizing the default value. 
For the shared global abstraction, all agents on the map are collected along with agents' features from the local observations, in addition to the weapon cooldown and previous action of each agent in the game. 
All features used in our experiments are normalized to $[0, 1]$ for both the local and abstracted global observation spaces.
Fig.~\ref{fig:sightRange} shows the default sight and shooting range of an individual agent in a SMAC environment.
In SMAC, the default achievable reward in a single episode is scaled to a value between 0 and 20 \cite{Samvelyan2019}, ignoring the negative rewards by default.
The reward is distributed to the entire team and takes into account factors such as inflicted damage on enemy units, points earned from killing units, and a bonus for achieving victory, rather than being individually assigned to each agent. Equation \ref{rewardEq} shows the shared reward computation for a full game episode,
\begin{equation}
    R=20\times\frac{\sum_{t=1}^{T}\left(\sum_{n=1}^{N}{\left(D_n\right)\times10\times k}\right)+200\times w}{R_{max}}
    \label{rewardEq}
\end{equation}
where $t$ is the current environmental step, $T$ is the terminal step, $n$ is the agent identifier, $N$ is the maximum number of agents, $D_n$ is the damage dealt to enemy units by agent $n$ in step $t$, $k$ is the number of enemies that have been defeated in step $t$, $R_{max}$ is the maximum possible reward amount, and $w$ is 1 on a win and 0 on a loss.
The episode reward is calculated when the game is over and a discounted reward with a decay of $0.9$ is used for training our RL models.

\subsection{Experimental Setup} 
Our MARL policies were evaluated on various SMAC scenarios, each with $31$ different random seeds for statistical analysis. After tuning the parameters representing the unified input and output, each of the experiments was conducted for $2000$ episodes. 
For statistical analysis, we evaluate the proposed model on $31$ random seeds in different SMAC scenarios. Each experiment runs for a total number of $2000$ episodes considering the large number of tuning parameters introduced in our neural network architecture. We built our neural network on top of the standard Advantage Actor-Critic (A2C) algorithm for both actor and critic networks \cite{9893711} utilized $\varepsilon$-soft, which combines the greedy and softmax methods with an initial value of 1 and gradually decreases as the process continues.
Equation \ref{epsilonEq} is used to calculate $\varepsilon$ at a specific time step where $\varepsilon_0$ is 0.0001 and the total number of environmental steps denoted by $\gamma$ is $30,000$.
\begin{equation}
    \varepsilon_t = max (\varepsilon_0 - \frac{t \times \varepsilon_0}{\gamma}, \varepsilon_{min})
    \label{epsilonEq}
\end{equation}

We use $\varepsilon$-soft instead of $\varepsilon$-greedy in order to select an action according to the softmax behavioral policy, which randomly selects an action, where the likelihood of selecting an action is the proportion of the action’s estimated value relative to the sum of the estimated values for all actions. This methodology was chosen because it promotes continued intelligent exploration of the state space, even when $\varepsilon$ itself becomes arbitrarily small. Exponential Linear Unit (ELU) has been used for the activation function where the value for $\alpha$ is 1. The learning rate used for each of the episodes is $0.0001$. 

\begin{figure*}[h]
    \centering    \includegraphics[width=1\textwidth]{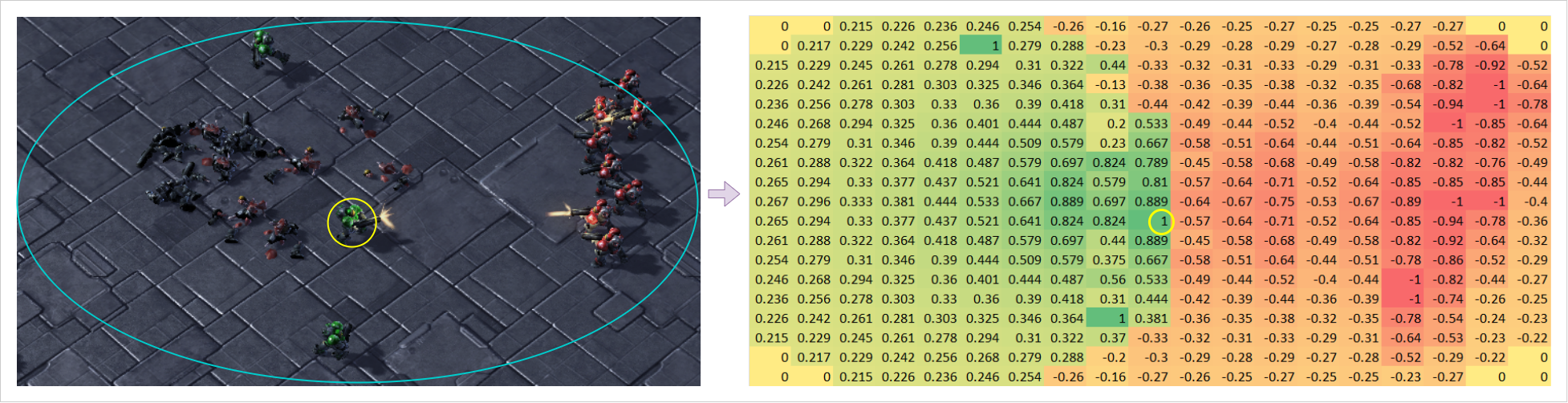}
    \caption{A Sample $19\times19$ Heatmap Generated from Local Observation on $8m$}
    \label{fig:localMAIM}
    \vspace{-4mm}
\end{figure*}

\subsection{Transfer Learning Architecture}
Reusing existing knowledge can greatly speed up the MARL learning process and make complex tasks more achievable for agents in MAS. This involves mapping a knowledge space to a policy, where the knowledge space is composed of samples from the current task, previous tasks, and other agents. 
To acquire an optimal policy, model-free MARL techniques require a considerable number of samples to be trained. It is challenging to learn optimal policies due to the large state and action spaces. For instance, there are various situations involving distinct units and terrain conditions in SMAC, and finding efficient strategies for each scenario from scratch can be time-consuming. 
Creating a generalized representation for the states independent of specific scenarios is a big challenge in SMAC scenarios. In the default state representation provided by SMAC, the states are represented based on the number of units in each scenario shown in Fig.~\ref{fig:transferLearningArc}a. 
For scenarios $3m$, $8m$, and $25m$, the size of local state is provided in the form of $30$, $80$, and $250$ one-dimensional vectors respectively which is marked with a dotted rectangle of red color in Fig.~\ref{fig:transferLearningArc}a. 
The default state representation is compact and carries precise agent information for training MARL. However, it limits the knowledge transfer over multiple scenarios due to the scenario-dependent state size.
To mitigate this problem, we further focused on constructing a fixed-size state representation that is unified in such a way that the state size remains constant, irrespective of the number of agents in the scenario. 

\subsubsection{Scenario Independent State Representation}\label{subsec:state}
Researchers have been experimenting with different approaches for unifying state representation to allow knowledge transferring in MARL. However, there is no scenario-independent state representation that has been widely accepted yet. 
In our approach, we consider two types of state information to feed into the reusable neural networks over multiple scenarios in SMAC, one is local observations, and the other is abstracted global information. 
In our prior work, we employed a spatial information technique called Agent Influence Map (AIM) to extract and filter aggregated spatial representation from the global information in order to discover common objectives and promote the learning of collaborative behaviors among agents. 
We further extended the use of AIM in this study to construct a scenario-independent local state representation for enabling knowledge transfer across all scenarios provided in SMAC.
Each AIM is determined by three parameters: the current relative health of the agent $I_0$, the influence decay rate which equals to the inverse of the distance from the agent $\lambda_I$, and the range of influence $d_I$.
A negative weight is used for enemy agents in order to differentiate from the allied agents.
With an aggregation of all the agents' AIM, a generalized and more robust Multi-Agent Influence Map (MAIM) is formed. Based on the performance of different dimensions, $64\times64$ MAIM representation has been used for unifying the abstracted global information in further experimentation.

In our state representation of local observation, we considered the local observations of the prior step, the actions performed in that step, along with the current step information. The local observations include distance, relative position, health, shield, and unit type for allied and enemy units within the sight range as demonstrated in Fig.~\ref{fig:sightRange}. The default states received from SMAC depend on the number of active agents in the game environment. 
In order to remove the dependency on the number of agents across SMAC scenarios, we extended the use of IM from global information abstraction to local observation aggregation.
The local IM transformation yields a fixed dimension of sight range with different resolutions such as $19 \times 19$, $37 \times 37$, and $55\times55$, each with the same local state parameters.
Fig.~\ref{fig:localMAIM} shows a sample heatmap with cell values transformed from a local state observation in a two-dimensional $19 \times 19$ matrix on $8m$ scenario. {This matrix is derived from the perspective of the highlighted agent where $1$ denotes the allies, and $-1$ indicates the enemy units.} 
The input dimension is determined by the agent's sight range in all four directions, resulting in a unified two-dimensional scaled matrix transformed from a local observation. In SMAC, the agents' sight range is a fixed value of 9, which means the agent can see an enemy within a distance of 9 cells. 
The unified subset of global and local information as shown in the green-colored rectangle in Fig.~\ref{fig:transferLearningArc}b is then flattened and propagate through the neural network, which applies to all SMAC scenarios.  
To unify the local observation using IMs without losing essential information, we evaluate different resolutions to construct our local IMs for various levels of granularity on unit manipulations during combats in SMAC. The generated IMs will then serve as scenario-independent input parameters for the neural network displayed in Fig.~\ref{fig:transferLearningArc}c, which carries the shared knowledge across all SMAC scenarios.
This unified representation of the local states resolves the dependency of a different number of agents in the environment, and the input of fixed dimensional local state, along with the abstracted global information from MAIM, is trained via neural networks to execute the probable attacks or actions for winning strategies in SMAC. 

\subsubsection{Scenario Independent Action Representation}
In SMAC, agents are required to make decisions choosing from a finite action space based on the state information described in the previous subsection. For the move actions, agents typically have four directions to choose from: north, south, east, and west. However, when it comes to attacking decisions, agents must consider the number of enemies in their current local observation within the sight range in the default action space provided in SMAC. This presents a challenge, as the number of enemies can vary greatly from scenario to scenario. To address this issue, we propose a generalized approach that only considers the closest enemy position for the attacking action to remove the dependency on the number of agents within the sight range and make attack actions without knowing specific enemy agents.
This approach is illustrated in Fig.~\ref{fig:transferLearningArc}d, where the scenario-dependent output is replaced by an attack action following our custom-generated policy.
As we trained our NNs with the spatial information of the agent's current position with detailed policies, targeting the closest enemy instead of choosing the default scenario-dependent action doesn't lose any valuable information required to take action. 
The generalized outcome enables the agent to share attacking policies, thereby facilitating the transfer of knowledge across a broad range of SMAC environments.
By reformulating the existing solution in a unified
manner, we were able to achieve both improved performance in large-scale scenarios and enhanced efficiency in the simulation environment.

\subsection{Curriculum Transfer Learning}
\begin{figure}[tb]
    \centering
    \includegraphics[width=0.45\textwidth]{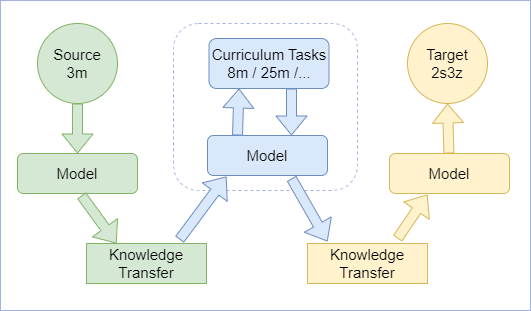}
    \caption{Curriculum Transfer Learning Architecture}
    \label{fig:CTL}
\end{figure}

Curriculum learning is a specific type of transfer learning that arranges a series of tasks according to their increasing level of complexity \cite{bengio2009curriculum}. 
This approach involved training on how to play a game against simulated opponents who became progressively more competent, allowing the agents to learn useful strategies and gain knowledge that they could apply to real-world challenges \cite{gupta2017insights}. 
We evaluated the performance of Curriculum Transfer Learning (CTL), which merges curriculum learning and transfer learning to facilitate faster convergence of the learning process and lead to better optimum outcomes {within the fixed time steps}.
We delve into the intriguing question of how the transfer of winning strategies among Marines learned in scenarios $3m$ and $8m$ can positively impact the behavior of Stalkers and Zealots in an extended heterogeneous environment $2s3z$. 
Fig.~\ref{fig:CTL} shows the curricular process flow of training our policy from the simplest scenario, $3m$, {retrain the learned model in a medium-level scenario}, $8m$, and finally carries the knowledge learned from $3m$ and $8m$ to tackle a much more complex scenario $2s3z$ with different unit types and heterogeneous team units that the policy has never seen in prior training scenarios. 

\section{Results and Discussion}\label{results} 
Transfer learning offers a potential solution to improve agents' asymptotic performance by enabling them to attain higher levels of performance in learning complex tasks in MAS.
The performance of our proposed TL approach has been evaluated on both homogeneous and heterogeneous scenarios in SMAC research platform through several performance metrics, such as the maximum, minimum, and average episode reward for all 31 running instances of each scenario.

\begin{table}[tb]
\caption{Best Performing RL Models Learning from Scratch}
\begin{center}
\begin{tabular}{c c c c c c}
\hline
\textbf{Scenario} & \textbf{Dimension$^{\mathrm{a}}$} & \textbf{Min} & \textbf{Max} & \textbf{Avg} & \textbf{Std}\\
\hline
$3m$ & $19\times19$ & 2.32 & 16.57 & 8.34 & \textbf{3.24} \\
& $37\times37$ & \textbf{4.29} & \textbf{20} & \textbf{12.39} & 5.21 \\
& $55\times55$ & 1.29 & 15.21 & 12.37 & 3.26 \\
\hline
$8m$ & $19\times19$ & 4.61 & 19.57 & 9.29 & 4.93 \\
& $37\times37$ & \textbf{8.19} & \textbf{20} & \textbf{11.74} & \textbf{3.34} \\
& $55\times55$ & 6.45 & 18.3 & 8.74 & 5.22 \\
\hline
$25m$ & $19\times19$ & 3.36 & 14.51 & 6.42 & 2.67 \\
& $37\times37$ & 4.93 & \textbf{15.97} & \textbf{7.45} & \textbf{1.28} \\
& $55\times55$ & \textbf{5.16} & 11.90 & 8.32 & 1.96 \\
\hline
\multicolumn{6}{l}{$^{\mathrm{a}}$ the dimension of unified local observation.}
\end{tabular}
\label{tab:seedPrep}
\end{center}
\vspace{-6mm}
\end{table}

\subsection{Pre-trained Policy Selection}
In the first step, we trained each of our selected homogeneous scenarios for 2000 epochs and 31 independent instances in parallel. The agents exploit the information from the ongoing simulation and train the neural networks accordingly. Performance comparison of these models was carried out based on different input state resolutions. Table~\ref{tab:seedPrep} represents the outcomes of $3m$, $8m$, and $25m$ with varying local state dimensions ranging from $19\times19$ to $55\times55$ resolutions. Results showed that the agents for $3m$ and $8m$ achieved a score of $20$ {which represents the agent successfully defeated their opponent team in the given scenario}. However, for $25m$, the highest score obtained by one of the best-trained models was $15.97$, with a  $37\times37$ dimension of local state representations. Upon considering the overall stability of the performance, the $37\times37$ dimension was chosen for local state reformulation in a unified manner. Consequently, the best-performing RL model from each of the $3m$, $8m$, and $25m$ scenarios is selected as a seed for training our shared neural networks on other scenarios, utilizing the transfer learning technique.

\begin{figure}[tb]
    \centering
        \subfloat[Average of the Running Average Episode Reward on $3m$\label{TL3m}]{\includegraphics[width=0.48\textwidth]{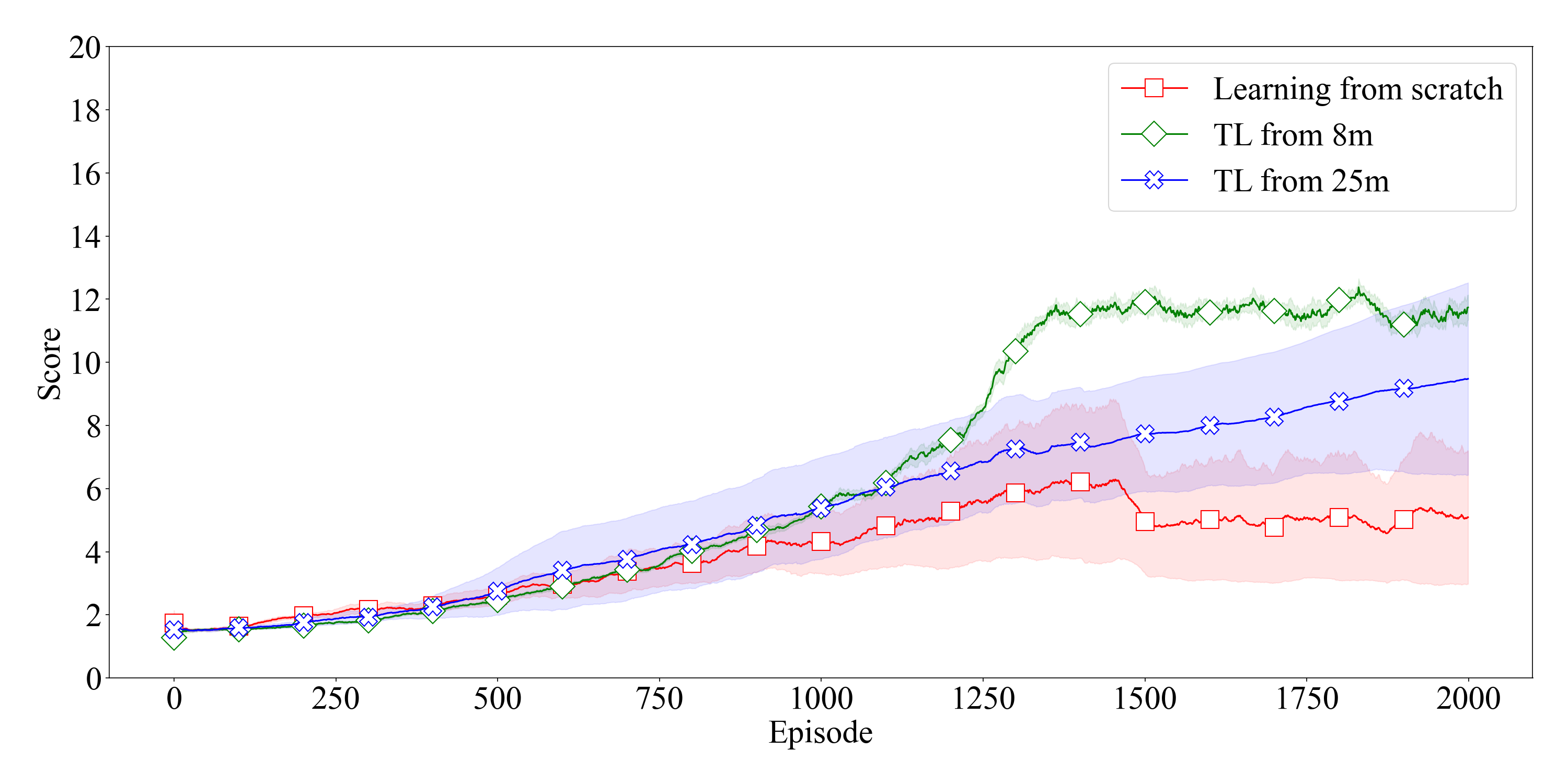}} \\
        \subfloat[Average of the Running Average Episode Reward on $8m$\label{TL8m}]{\includegraphics[width=0.48\textwidth, height=4cm]{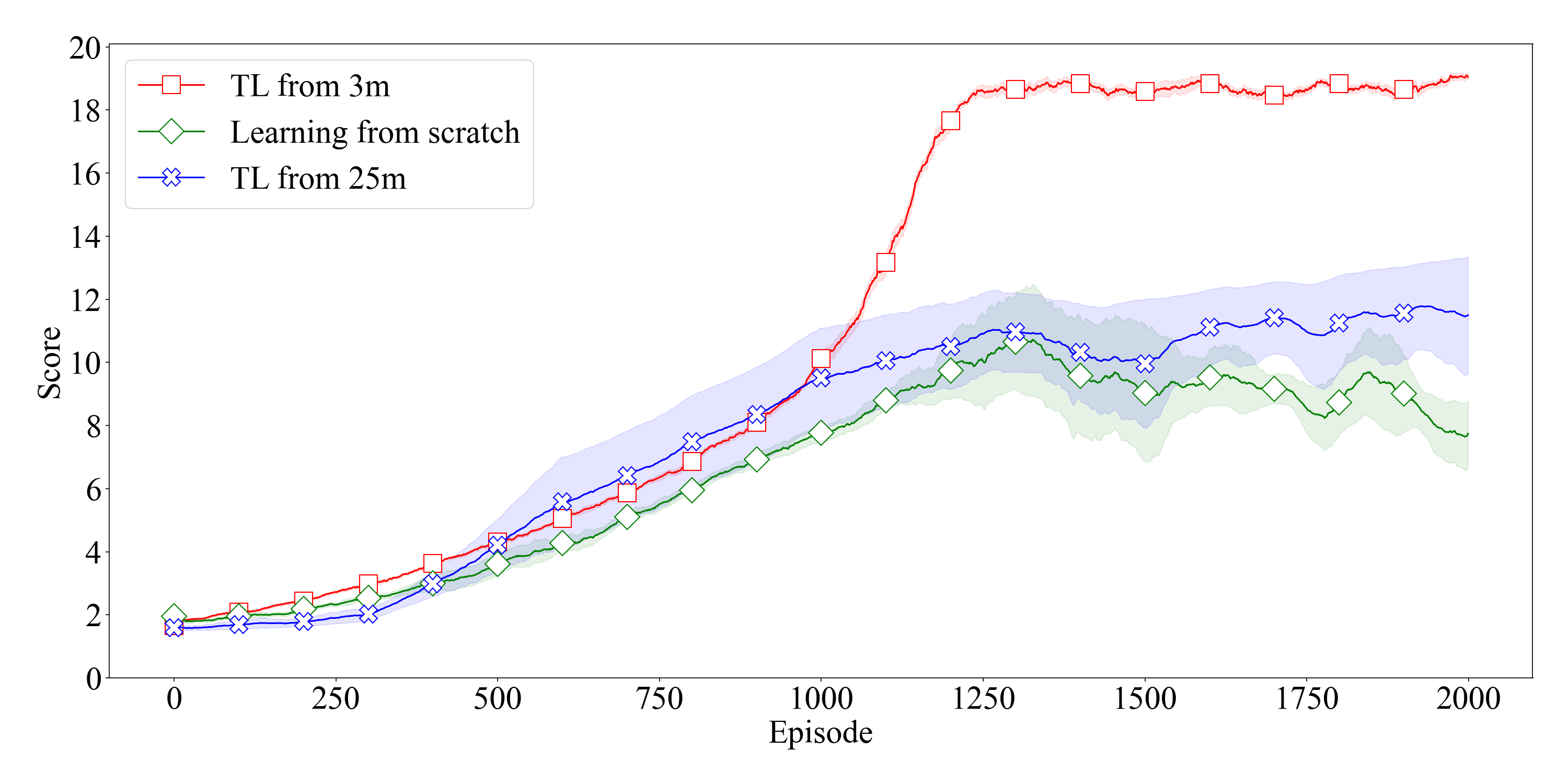}} \\
        \subfloat[Average of the Running Average Episode Reward on $25m$\label{TL25m}]{\includegraphics[width=0.48\textwidth, height=4cm]{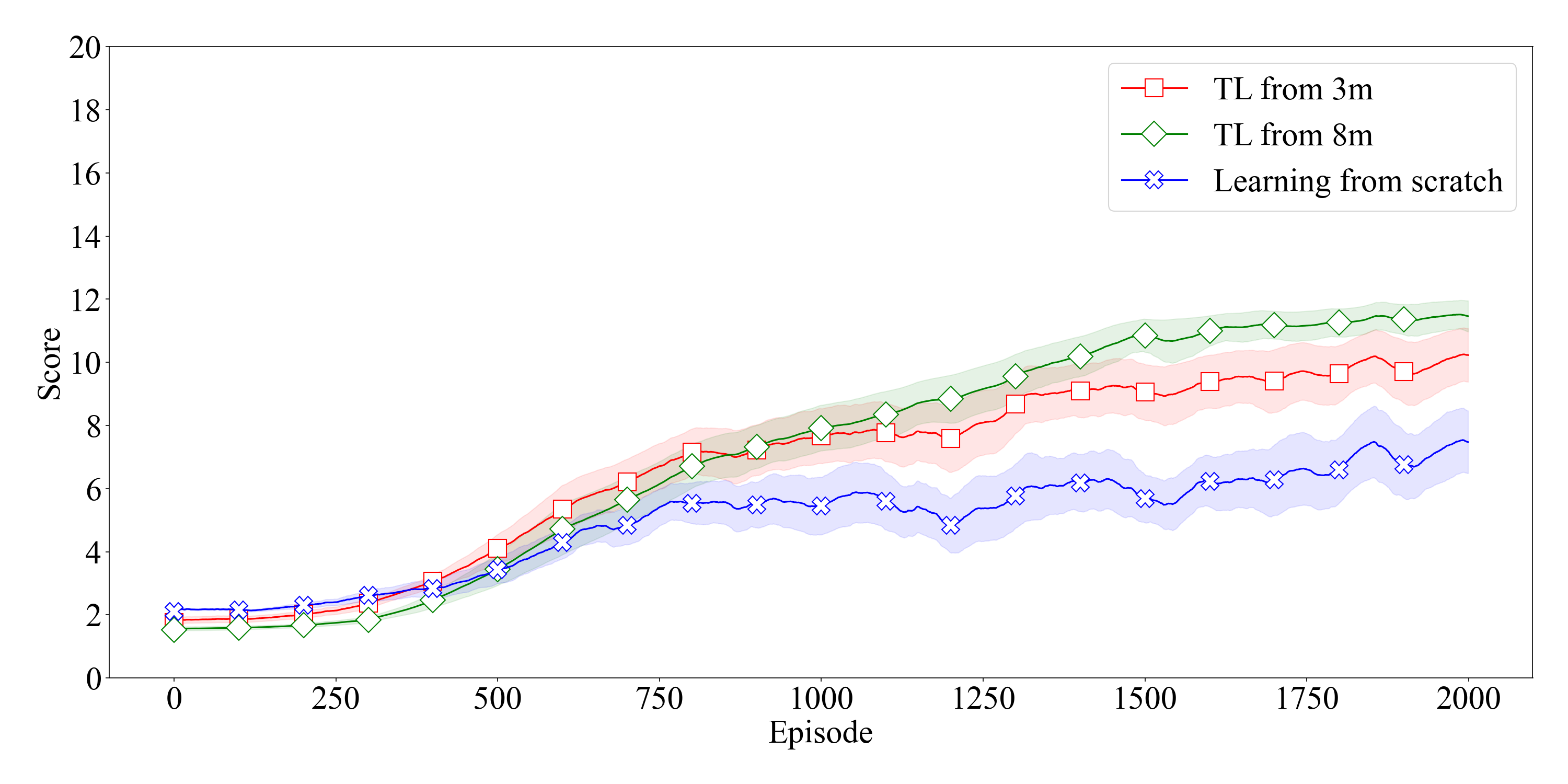}}
    \caption{Results of Transfer Learning on $3m$, $8m$, and $25m$}
    \label{TLStatGraph}
    \vspace{-6mm}
\end{figure}

\subsection{Transfer Learning Performance}
The seeds selected from different homogeneous scenarios have been utilized for training other scenarios to evaluate the performance of transferring knowledge across scenarios in SMAC.
Fig.~\ref{TLStatGraph} illustrates the performance improvement in $3m$, $8m$, and $25m$ scenarios using TL based on different seed policies.
In Fig.~\ref{TL3m}, we can see that the performance of $3m$ is improved significantly when the knowledge of $8m$ and $25m$ is used instead of learning from the beginning without any shared information. The average running episode reward of $3m$ is enhanced by $9.7\%$ and $19.97\%$ when the pre-trained policy of $8m$ and $25m$ is utilized, respectively. The reason behind the improvement of decision making is that both $8m$ and $25m$ have a larger number of Marines actively learning, and extracting those information boosts the learning performance compared to $3m$ where only three active Marines are fighting against the enemy units. 
Fig.~\ref{TL8m} reflects the performance of $8m$ on the scale of maximum average running episode reward. The performance curve of TL-$3m$ depicts the impact of the seed policy $3m$ on $8m$ scenario with an overall $45.3\%$ average score improvement. We can see that the pre-trained models performed poorly at the beginning of the training due to the change in the unseen scenario and then outperformed other models after 1000 iterations, proceeding toward the winning strategy in each episode afterward. It took However, reusing the pre-trained $25m$ model boosts the average performance with an overall improvement of $8.01\%$.
Though our model couldn't ensure stable winning for $25m$, it tweaked with a better representation to deal with enemy units and slightly surpassed the base units' performance.
Fig.~\ref{TL25m} shows the impact of seeds $3m$ and $8m$ on the $25m$ scenario, where the average running episode reward is enhanced by $36.4\%$ and $41.3\%$ respectively compared to the learning from scratch approach. 

Table~\ref{transferLearningStats} reflects the statistical analysis of the TL model on extended homogeneous SMAC environment. To examine the effectiveness of our proposed approach, four primary metrics are considered: the maximum, minimum, average, and standard deviation (STD) of the average of the running average episode reward with respect to various thresholds. All these evaluations are performed across a complete set of 31 seeds for statistical analysis.
The highest minimum average running episode reward score is $5.21$, $11.5$, and $6.54$ for $3m$, $8m$, and $25m$, respectively, when the pre-trained seed policies $8m$, $3m$, and $8m$ are incorporated in the training phase instead of learning from scratch. Similarly, the maximum and average of the running episode outperformed the base scenarios due to the effect of TL and taking action based on the knowledge of other scenarios. The fluctuation in the score is minimized, depicted by the STD value. The smaller the value of STD, the more stable the model is in terms of performance. The boldly marked values are the highest values highlighted in different SMAC scenarios.

\begin{table}[tb]
\caption{Performance Evaluation of TL in SMAC scenarios}
\begin{center}
\begin{tabular}{c c c c c c}
\hline
\textbf{Scenario} & \textbf{Pretrained Policy} & \textbf{Min} & \textbf{Max} & \textbf{Avg} & \textbf{Std}\\
\hline
$3m$ & $-$ & 4.29 & 18.34 & 12.37 & 3.74 \\
& $8m$ & \textbf{5.21} & \textbf{20} & 13.57 & \textbf{0.28} \\
& $25m$ & 3.31 & 17.78 & \textbf{14.84} & 3.12 \\
\hline
$8m$ & $3m$ & \textbf{11.5} & \textbf{20} & \textbf{17.06} & \textbf{0.1} \\
& $-$ & 8.19 & 18.41 & 11.74 & 3.34 \\
& $25m$ & 10.95 & 17.8 & 12.68 & 1.89 \\
\hline
$25m$ & $3m$ & 4.31 & 13.59 & 11.35 & 1.69 \\
& $8m$ & \textbf{6.54} & \textbf{14.39} & \textbf{11.76} & \textbf{1.28} \\
& $-$ & 5.16 & 11.9 & 8.32 & 1.96 \\
\hline
\multicolumn{6}{l}{$-$ indicates learning from scratch.}
\end{tabular}
\label{transferLearningStats}
\end{center}
\vspace{-4mm}
\end{table}

\begin{figure}[tb]
    \centering
    \includegraphics[width=0.48\textwidth]{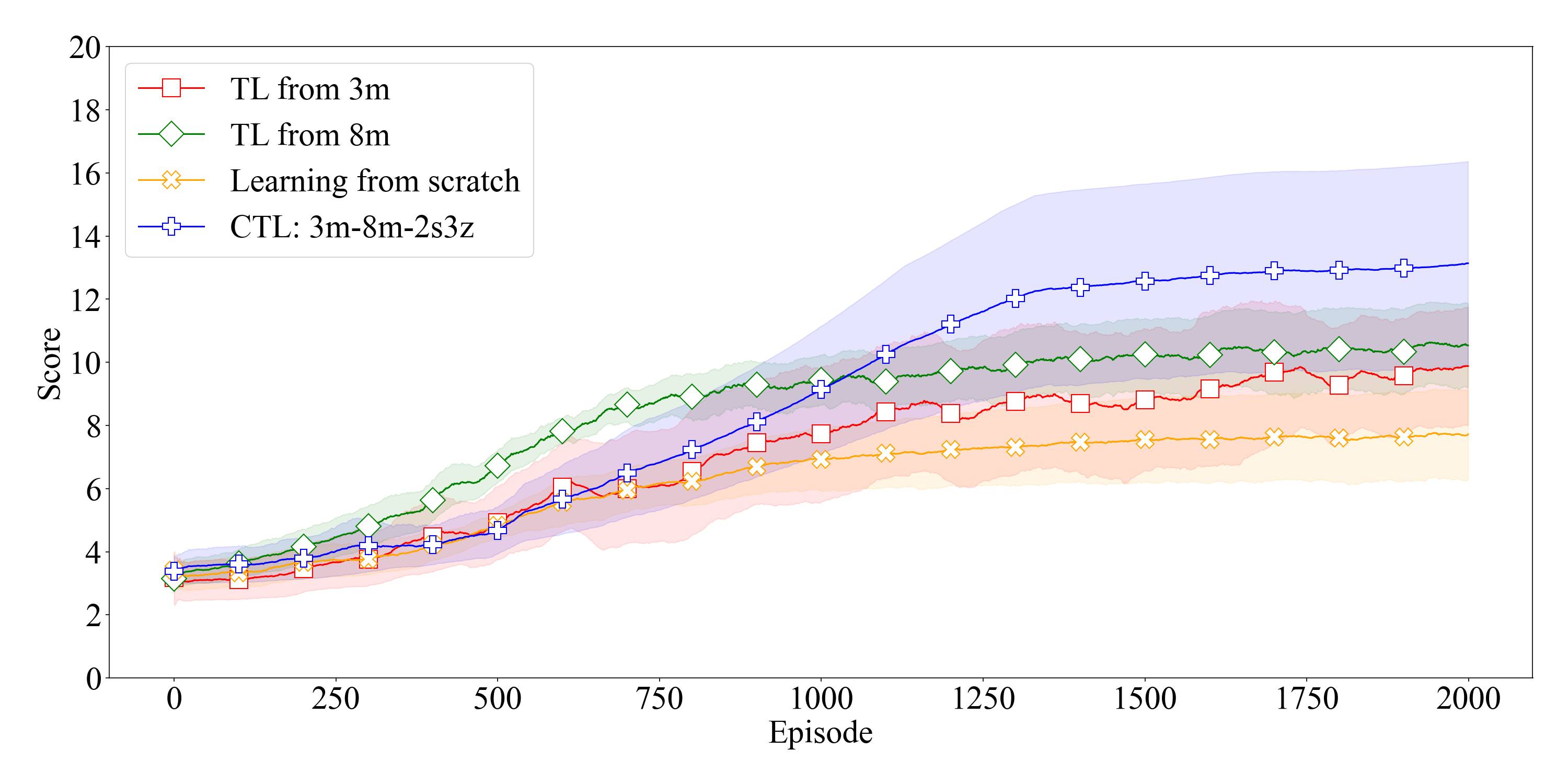}
    \caption{Result of Curriculum Transfer Learning on $2s3z$}
    \label{fig:ctl2s3z}
    \vspace{-4mm}
\end{figure}

\subsection{Curriculum Transfer Learning Performance}
As of yet, our study investigates the impact of TL on homogeneous scenarios for evaluating intra-agent knowledge transfer, and the experimental results demonstrate a significant improvement in MARL performance. 
We further aim to extend our evaluation to complex heterogeneous scenarios evaluating not only the intra-agent knowledge transfer within the same unit type, but also the inter-agent knowledge transfer among different types of units. To this end, we have selected the $2s3z$ testing bed and compared the performance using seeds from prior experiments. We scrutinized the outcome of CTL in which the $3m$ model was trained at first, as this is one of the simplest maps in SMAC. After training each $3m$ instance for $2000$ episodes, we saved the best-performing model and utilized it for $8m$, which is comparatively difficult to train. The knowledge learned to control Marines from $3m$ and $8m$ is then carried out for the most challenging scenario of our test environment, $2s3z$ containing Stalkers and Zealots that the model has never seen in the previous training scenarios. 

\begin{table}[tb]
\vspace{-4mm}
\caption{Performance Evaluation of CTL on $2s3z$}
\begin{center}
\begin{tabular}{c c c c c c}
\hline
\textbf{Scenario} & \textbf{Pretrained Seed} & \textbf{Min} & \textbf{Max} & \textbf{Avg} & \textbf{Std}\\
\hline
$2s3z$ & $3m$ & 5.85 & 14.79 & 11.13 & 3.29 \\
& $8m$ & \textbf{8.91} & 16.19 & 11.1 & \textbf{3.24} \\
& $-$ & 5.68 & 13.61 & 8.03 & 3.66 \\
& CTL$^{\mathrm{a}}$ & 4.93 & \textbf{17.2} & \textbf{13.83} & 4.98 \\
\hline
\multicolumn{6}{l}{$-$ indicates learning from scratch.} \\
\multicolumn{6}{l}{$^{\mathrm{a}}$ CTL: $3m \rightarrow 8m \rightarrow 2s3z$} \\

\end{tabular}
\label{tab:CTLStats}
\end{center}

\end{table}

Fig.~\ref{fig:ctl2s3z} depicts the outcome of the average running episode reward for $2s3z$ model in different TL and CTL scenarios. The graph shows the upward trend of our curriculum model when the information is carried out from $3m$ to $8m$ and then $8m$ to $2s3z$. The TL model using pre-trained seeds $3m$ and $8m$ improved the average reward of $2s3z$ by $38.6\%$ and $38.2\%$ compared to the performance of learning from scratch on $2s3z$. The performance of all these simulations was surpassed by the performance of CTL with an overall improvement of $72.2\%$.
Table~\ref{tab:CTLStats} demonstrates the statistical representation of the performance metrics. The curriculum transfer $3m\rightarrow8m\rightarrow2s3z$ works best with the maximum average reward of $17.2$ and the highest average of $13.83$ in our simulation environment. This ensures our curriculum architectures' robustness in intra- and inter-agent knowledge transfer. 

\section{Conclusion and Future Work}\label{conclusion}
This study introduced a novel approach to unifying various state sizes into fixed-size inputs to enable transfer learning and curriculum transfer learning in different MARL systems.
One of the challenges in designing effective MAS is the need for agents to learn from limited data and quickly adapt to new environments. Transfer learning allows agents to leverage knowledge gained from previous experiences to accelerate learning in new domains, reducing the need for extensive data collection and training.
We evaluated the performance of our TL and CTL using the uniformed state and action representation in both homogeneous and heterogeneous SMAC scenarios with varying complexity. By pre-training agents on similar tasks, they were able to develop a better understanding of the underlying concepts and principles needed to perform new, complex tasks. We observed significant improvements in overall learning performance, generalizability, and peak performance compared to our previous approach. Our findings could inspire the development of better-unified representations and feature discovery in MARL algorithms.

There are several areas can be further explored from this study. Firstly, our model was tested on a limited number of scenarios, and future research could investigate its performance in a wider range of heterogeneous environments with more complex maps. Additionally, we can explore the use of multiple influence maps and other kernel functions to discover correlations between spatial features at local, shared, and global levels. Finally, integrating other deep RL techniques, such as recurrent neural networks, could expand the possibilities for improvement in multi-agent systems.

\section{Acknowledgement}
This material is based upon work supported by the National Science Foundation under Grant OAC-1828380.

\bibliographystyle{IEEEtran}
\bibliography{References}

\end{document}